\documentclass[10pt,twocolumn,letterpaper]{article}

\usepackage[final]{cvpr}      %
\usepackage[utf8]{inputenc} %
\usepackage[T1]{fontenc}    %
\usepackage[pagebackref,breaklinks,colorlinks]{hyperref}
\usepackage{url}            %
\usepackage{booktabs}       %
\usepackage{floatrow}
\usepackage{amsfonts}       %
\usepackage{amsmath}
\usepackage{nicefrac}       %
\usepackage{microtype}      %
\usepackage[dvipsnames,table,xcdraw]{xcolor}         %
\usepackage{graphicx}  %
\usepackage{paralist}
\usepackage{makecell}
\usepackage{adjustbox}
\usepackage{etoc}
\usepackage{multirow}
\usepackage{colortbl}
\usepackage{pifont}
\usepackage{xspace}
\usepackage{rotating}
\usepackage{bbm}
\usepackage{subcaption}
\usepackage[font=small]{caption}
\def\etal{\emph{et al}.\xspace}

\newfloatcommand{capbtabbox}{table}
\newfloatcommand{capbfigbox}{figure}

\definecolor{Gray}{gray}{0.9}

\newcolumntype{R}[2]{%
    >{\adjustbox{angle=#1,lap=\width-(#2)}\bgroup}%
    l%
    <{\egroup}%
}
\newcommand*\rotb{\multicolumn{1}{R{45}{1em}}}%
\newcommand{\method}{\texttt{AUGCO}\xspace}
\newcommand{\source}{\mathcal{S}}  %
\newcommand{\target}{\mathcal{T}}  %
\newcommand{\numsamples}{N}
\newcommand{\model}{h}  %

\DeclareMathOperator*{\argmax}{argmax}

\newcommand*{\TODO}{\textcolor{black}}

\def\cvprPaperID{1781} %
\def\confName{CVPR}
\def\confYear{2022}
\title{\method: Augmentation Consistency-guided Self-training for Source-free Domain Adaptive Semantic Segmentation}

\author{
    \textbf{Viraj Prabhu}$^{*}$ \qquad
    \textbf{Shivam Khare}\thanks{Equal contribution} \qquad
    \textbf{Deeksha Kartik} \qquad 
    \textbf{Judy Hoffman} \qquad \\
    Georgia Institute of Technology \\
    {\small\texttt{\{virajp,skhare31,dkartik3,judy\}@gatech.edu}}    
}

\begin{document}

\maketitle

\begin{abstract}
  Most modern approaches for domain adaptive semantic segmentation rely on continued access to source data during adaptation, which may be infeasible due to computational or privacy constraints. We focus on source-free domain adaptation for semantic segmentation, wherein a source model must adapt itself to a new target domain given only unlabeled target data. We propose Augmentation Consistency-guided Self-training (\method), a  source-free adaptation algorithm that uses the model's pixel-level predictive consistency across diverse, automatically generated views of each target image along with model confidence to identify reliable pixel predictions, and selectively self-trains on those. 
  \method achieves state-of-the-art results for source-free adaptation on 3 standard benchmarks for semantic segmentation, all within a simple to implement and fast to converge method.
\end{abstract}

  \section{Introduction}
  \label{sec:intro}

  In this work, we focus on the challenging problem of \emph{source-free} domain adaptation~\cite{liang2020we,li2020model,kim2020domain,kundu2020universal} for semantic segmentation~\cite{wang2021tent,liu2021sourcefree}. 
  Consider a deep model trained to perform semantic segmentation deployed atop an autonomous vehicle. While unsupervised domain adaptation (DA) has been extensively studied~\cite{saenko2010adapting,ganin2014unsupervised,long2015learning,hoffman2017cycada,vu2019advent}, most prior DA methods assume continued access to labeled source data during adaptation. In our example, this may be impractical due to the limitations of on-board compute and memory, particularly so for a compute-heavy task such as segmentation. Further, such access to source data on-board may also be subject to and limited by privacy regulations.
    
  Concretely, our goal is to adapt a trained semantic segmentation model to a new target domain given only its trained parameters and unlabeled target data. 
  The absence of source data for regularization makes this source-free adaptation setting very challenging and highly susceptible to divergence from original task training, leading to catastrophic performance loss.

  To address the challenging nature of this task, recent methods have introduced complex multi-part
  solutions: Liu~\emph{et al.}~\cite{liu2021sourcefree} propose an approach combining attention, knowledge distillation, self-training, and patch-level self-supervised learning, whereas Fleuret~\emph{et al.}~\cite{fleuret2021uncertainty} propose combining entropy regularization with feature corruption using multiple auxiliary decoders. These methods introduce several hyperparameters which make them challenging to tune, particularly so in the absence of any labeled data whatsoever. 
    
  \begin{figure}[t]
    \centering
    \includegraphics[width=0.9\textwidth]{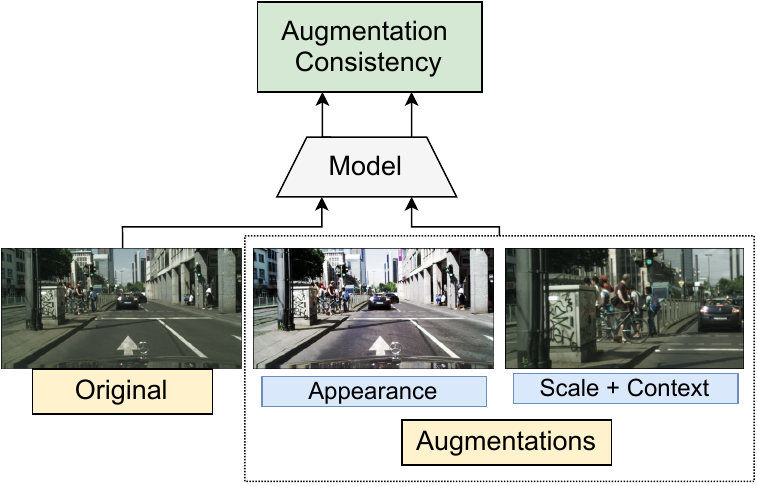}
    \caption{We study source-free domain adaptive semantic segmentation, where the goal is to adapt a well-trained source model to a target domain given only unlabeled target data. Under a domain shift, many source model predictions on the target domain are initially incorrect; unconstrained self-training would reinforce such errors and degrade performance. We propose \method, a selective self-training algorithm that identifies reliable predictions on which to self-train based on pixel-level predictive consistency across diverse target image views that vary in appearance, scale, and context, and leads to state-of-the art performance in the source-free setting.}
    \label{fig:teaser}
  \end{figure}

  In contrast, an alternative line of work has focused on a remarkably simple solution: parameter constrained self-training. For example, Test-time adaptation by entropy minimization or TENT~\cite{wang2021tent}, constrains optimization to only update the model's batch-norm parameters (both affine and normalization), and self-trains on unlabeled target data by minimizing a conditional entropy~\cite{grandvalet2005semi} loss. By keeping all other parameters frozen, TENT is able to prevent task drift in the source-free adaptation setting.
    
  While TENT leads to modest performance improvements on standard domain shifts, it performs self-training on \emph{all} model predictions. Under a domain shift, many of the model's predictions may initially be incorrect, and entropy minimization encourages the model to increase its confidence even on such incorrect predictions! As a result, unconstrained self-training leads to error accumulation~\cite{chen2019progressive,jiang2020implicit,prabhu2020sentry}, particularly on categories on which the source model does poorly to begin with.
    
  To address this, prior work has proposed \emph{selective} self-training on instances deemed \emph{reliable} via model confidence~\cite{tan2019generalized} or consistency under random image augmentations~\cite{prabhu2020sentry}. However, model confidence from deep networks is known to be miscalibrated under a domain shift~\cite{snoek2019can}, and the suitability of augmentation consistency for semantic segmentation has not been  previously studied. We propose a novel selection strategy that combines pixel-level predictive consistency across diverse, automatically generated target image views with per-class confidence.
    
  Specifically, we generate two views of each target image that vary in scale, spatial context, and color statistics via a simple crop, resize, and color jiter strategy (see Fig.~\ref{fig:teaser}). We then obtain aligned model predictions in both views and mark pixels for which the model makes identical predictions in both views as ``reliable''. Note that this deviates from several recent works that propose \emph{encouraging} consistency between predictions across different views~\cite{chen2020simple,he2020momentum,xie2020propagate}; we instead propose \emph{measuring} this consistency to identify reliable pixels for self-training. Next, we also mark pixels in the top-K (K is a hyperparameter) percentile by confidence per-category as reliable. 
    
The model is then selectively self-trained on predicted pseudolabels for reliable predictions. To prevent task drift in the source-free setting, we match TENT~\cite{wang2021tent} to constrain weight updates to only the model's batch-norm parameters. We make the following contributions:

  \begin{enumerate}
    \item We propose Augmented Consistency-guided Self-training (\method), a simple source-free adaptation algorithm for semantic segmentation that identifies reliable pixel predictions by combining pixel-level predictive consistency across diverse, automatically generated target image views with model confidence, and then selectively self-trains on those.
    \item \method pushes the state-of-the-art on source-free adaptation from GTA5~\cite{richter2016playing}$\to$Cityscapes~\cite{cordts2016cityscapes} (+3.9 mIoU) and Cityscapes$\to$Dark Zurich~\cite{sakaridis2019guided} (+5.8 mIoU), and matches it on SYNTHIA~\cite{ros2016synthia}$\to$Cityscapes, with no extra parameters and within a single epoch of DA.
  \end{enumerate}
\section{Related Work}
\label{sec:relwork}
\noindent\textbf{Unsupervised domain adaptation (UDA) for semantic segmentation.} Initial approaches to UDA for semantic segmentation employed pixel-level adversarial learning~\cite{hoffman2016fcns}, adversarial learning in the output space~\cite{tsai2018learning}, or adversarial learning with feature purification~\cite{luo2019significance}. Follow-up work has incorporated pixel-level constraints such as  image-to-image translation~\cite{murez2018image}, cycle consistency and inconsistency~\cite{hoffman2018cycada,kang2020pixel} and several others~\cite{chen2019crdoco,fu2019geometry,yang2020fda,yang2020phase,chen2018road,zhang2019curriculum,zhang2019curriculum,zou2018unsupervised,pan2020unsupervised,sun2019not,li2020content}.
We propose a source-free UDA approach for semantic segmentation that does not require adversarial training, generative modeling, or cross-domain constraints, but instead makes use of self-supervised signals in the form of pixel-level predictive consistency across diverse target image views.

\noindent\textbf{Self-training for UDA.} Recently, training on model predictions or \emph{self-training} has been shown to be an effective UDA strategy~\cite{zou2019confidence,vu2019advent,prabhu2020sentry,wei2021theoretical}. Self-training typically involves simple supervised training on the model's predictions on unlabeled target data (\emph{pseudolabels}) to minimize a cross-entropy~\cite{tan2019generalized}, conditional entropy~\cite{vu2019advent,zou2018unsupervised} or max-squares objective~\cite{chen2020simple}, often with additional confidence regularization~\cite{zou2019confidence} or other constraints~\cite{lian2019constructing,mei2020instance,tang2021unsupervised}. 
However, unconstrained self-training leads to error accumulation under a domain shift. Recent work has sought to resolve this via prototype-based pseudolabel denoising~\cite{zhang2021prototypical}, self-training based on confidence~\cite{tan2019generalized}, or committee consistency~\cite{prabhu2020sentry}. We extend selective self-training to semantic segmentation and design a novel selection strategy that combines predictive consistency across aligned image views that differ in scale, context, and color statistics, with per-class  confidence.

\noindent\textbf{Source-free and test-time adaptation.} Recent work has studied the \emph{source-free} UDA setting, wherein a trained source model must adapt itself to a target given access only to unlabeled target data~\cite{liang2020we,li2020model,kim2020domain,kundu2020universal}. These works focus on image classification and propose approaches based on source hypothesis transfer~\cite{liang2020we}, class-conditional generative modeling~\cite{li2020model}, progressive pseudolabeling~\cite{kim2020domain}, and source-similarity based weighting~\cite{kundu2020universal}. Some recent works study source-free UDA for semantic segmentation: Liu~\etal~\cite{liu2021sourcefree} propose a solution based on data-free distillation, self-training, and patch-level self-supervision. Fleuret~\etal~\cite{fleuret2021uncertainty} combine self-training, entropy minimization, and feature noising. Kundu~\emph{et al.}~\cite{kundu2021generalize} recently propose an approach combining extensive domain generalization followed by adaptation. While source-free UDA typically adapts models to target train data and evaluates performance on a held-out target test set, the related task of test-time adaptation~\cite{sun2019test,wang2021tent} directly adapts model to target test data. 
We match the source-free UDA setting and propose a selective self-training approach that adapts the model in a single epoch of training.

\noindent\textbf{Batch normalization and domain adaptation.} Some works have focused on the role of batch normalization~\cite{ioffe2015batch} (BN) layers in adapting to new domains, proposing adaptive BN ~\cite{li2016revisiting} and domain-specific BN~\cite{chang2019domain}. Recent works propose test-time updates to BN parameters as a means of overcoming covariate shift without source data~\cite{nado2020evaluating,wang2021tent} -- Nado~\etal propose re-estimating BN normalization parameters on test data, whereas Wang~\etal propose TENT, which additionally updates BN channel-wise affine parameters over test/target data to minimize predictive entropy. Motivated by the success of these works in source-free adaptation and recent findings on the large expressive power of BN layers~\cite{frankle2021training}, we also constrain optimization to the network's batch-norm parameters during training.

\noindent\textbf{Predictive consistency.} Predictive consistency under image transformations has been successfully applied as a regularizer in supervised learning~\cite{cubuk2020randaugment}, self-supervised learning~\cite{chen2020simple,he2020momentum,xie2020propagate}, semi-supervised learning~\cite{sajjadi2016regularization,berthelot2019mixmatch,xie2020unsupervised,sohn2020fixmatch}, and unsupervised DA~\cite{li2020rethinking,wei2021theoretical,araslanov2021self}. Unlike prior work which \emph{optimizes} for consistent predictions across augmentations, we propose \emph{measuring} the model's pixel-level predictive consistency across diverse views of each target image to detect reliable predictions on which to self-train. Some prior work has considered such a measure for classification models: for error detection~\cite{bahat2019natural}, and standard UDA~\cite{prabhu2020sentry}, whereas we focus on source-free UDA for semantic segmentation.

\newcommand{\imgInputSpace}{\mathcal{X}}
\newcommand{\imgInput}{\mathbf{x}}
\newcommand{\imgInputTgt}{\imgInput_{\target}}
\newcommand{\imgInputTgtRescale}{\tilde{\imgInput}_{\target}}
\newcommand{\outputSpace}{\mathcal{Y}}
\newcommand{\outputLbl}{y}
\newcommand{\pseudoLbl}{\hat{\outputLbl}}
\newcommand{\numCls}{C}
\newcommand{\outputProb}{\mathbf{p}}

\newcommand{\Va}{V}
\newcommand{\Vb}{\tilde{V}}

\section{\method: Augmentation Consistency-guided Self-Training}
\label{sec:approach}

\begin{figure*}[t]
    \includegraphics[width=0.9\linewidth]{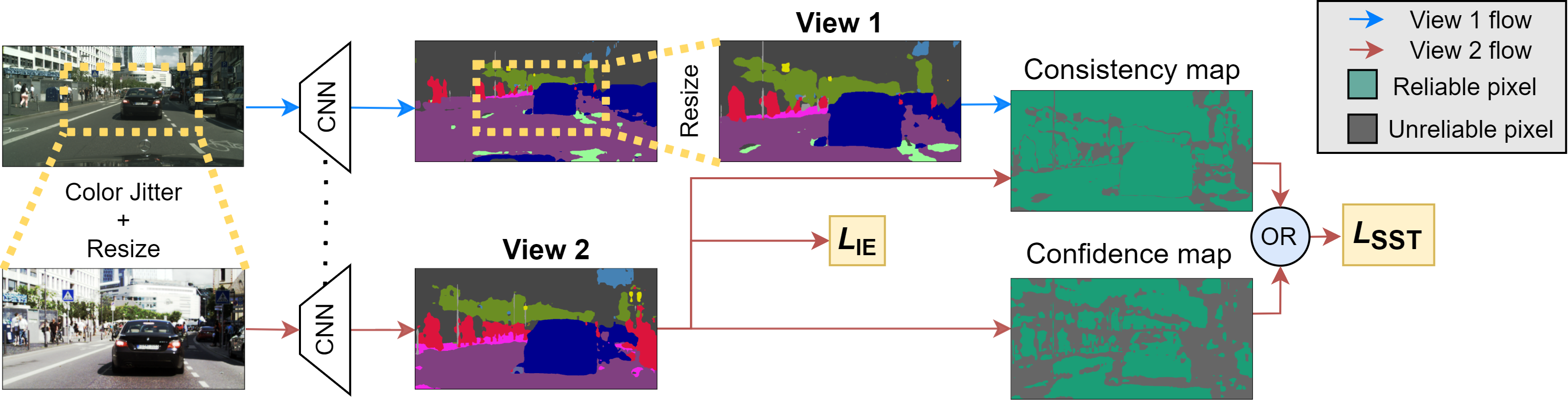}
    \caption{Overview of Augmentation Consistency-guided Self-Training (\method). \textbf{Left:} First, the model makes predictions on two views of each target image that differ in scale, spatial context and color statistics, that are generated via a random crop, resize, and jitter strategy (Sec.~\ref{sec:viewgen}). \textbf{Right:} Next, reliable pixel predictions for self-training are identified based on pixel-level consistency across aligned predictions and class-conditioned confidence thresholding, followed by selective self-training (Sec.~\ref{sec:sst}).}
    \label{fig:approach}
 \end{figure*}

\noindent \textbf{Setup and Notation.}
We study source-free domain adaptation for semantic segmentation. In semantic segmentation we are given an input image, $\imgInput \in \mathbb{R}^{H \times W \times 3}$, and the goal is to label every pixel, $\imgInput_{ij}$, with one of $\numCls$ semantic labels, $\outputLbl_{ij} \in \{1,2,\dots,\numCls\}$, producing an output label image, $\outputLbl \in \mathbb{R}^{H \times W}$.
To do this, we will learn a function $\model$ (CNN in our case) which takes images as input and produces a probabilistic output over $\numCls$ classes for each output pixel: $\model: \imgInput \rightarrow \outputProb \in \mathbb{R}^{H\times W\times \numCls}$. 
We produce a \emph{pseudolabel} by taking the argmax of the output probabilities: $\pseudoLbl = \argmax \outputProb$.
In source-free domain adaptation, we assume access to a model trained on labeled source ($\source$) data, $\model_\source$, as well as $\numsamples$ \emph{unlabeled} instances $\imgInput_\target \sim \mathcal{P}_\target(\imgInputSpace)$ from a target domain $\target$. 

\noindent\textbf{Overview.} The main goal of our algorithm is to learn a target model, $\model$, by leveraging a trained source model ($\model=\model_\source$ at initialization) and unlabeled target data. Standard unconstrained self-training under a domain shift suffers from the problem of error accumulation. To address this, we present \method, a selective self-training strategy for source-free domain adaptive semantic segmentation (see Fig.~\ref{fig:approach}). 

Our method first uses a random crop, resize, and jitter strategy to generate two aligned predictive views of each target image that capture model predictions across varying object scale, spatial context, and color statistics (Sec.~\ref{sec:viewgen}). Next, \method identifies \emph{reliable} model predictions on which to self-train using self-supervised signals in the form of pixel-level predictive consistency across the two aligned views, as well as model confidence. Finally, the model is self-trained using pseudolabels for reliable predictions (Sec.~\ref{sec:sst}).

\subsection{Aligned predictive view generation}
\label{sec:viewgen}

A key facet of our approach will be to identify pixels for which model predictions are deemed reliable. To do this we ensemble model predictions over random image regions that differ in scale and spatial context. We begin by randomly selecting a bounding box with coordinates, $(r_1,c_1, r_2, c_2)$, for each target image that satisfies two constraints: i) it spans an area that is 25-50\% of the area of the original image and ii) it matches the aspect ratio of the original image (i.e. $(r_2-r_1)/(c_2-c_1) = H/W$).

\noindent\textbf{View 1 (resized crop of prediction):} To create the first output prediction, we pass the original image, $\imgInputTgt$, through the current model, $\model$, to produce an output probabilistic prediction, $\outputProb = \model(\imgInputTgt)$. This original output prediction will be cropped using the random bounding box coordinates and resized to the original output image size: 
$\Va = \texttt{resize}(\argmax \outputProb[r_1:r_2, c_1:c_2], H, W)$

\noindent\textbf{View 2 (prediction on resized image crop):} For our second output prediction we first modify image appearance by applying a pixel-level color jitter $\imgInputTgt' = \texttt{jitter}(\imgInputTgt)$. We then use the same bounding box coordinates to extract a cropped image region and resize that region to the original image size to produce a rescaled image view $\imgInputTgtRescale = \texttt{resize}(\imgInputTgt'[r_1:r_2,c_1:c_2], H,W)$. This jittered, cropped, and resized image is then passed through the model to produce a probabilistic output, $\tilde{\outputProb} = \model(\imgInputTgtRescale)$ and associated predicted view, $\Vb = \argmax \tilde{\outputProb}$.

We thus obtain aligned predictive views $\Va$ and $\Vb$, which capture model predictions made at varying object scale (\emph{e.g.} in Fig.~\ref{fig:approach}, the size of the car in the secondary view is larger than in the original), spatial context (\emph{e.g.} additional cars are absent in the secondary view), and color statistics. 
We further refine predictions in both views via flip ensembling: predictions on a horizontally flipped version of each view are flipped back and averaged with the original predictions.

\subsection{Selective Self-Training}
\label{sec:sst}
\vspace{-3pt}

To increase the robustness of self-training, our method identifies pixel predictions on which to self-train using two self-supervised reliability signals: predictive consistency across aligned views and model confidence. The two sources offer complementary information about the model's stability across augmentation and its intrinsic confidence. 

\subsubsection{Measuring Reliability}

\noindent\textbf{Pixel-level predictive consistency.} First, we measure pixel-level consistency between the model's aligned predictions $\Va$ and $\Vb$,  %
and mark pixels with identical predictions ($\Va_{ij} == \Vb_{ij}$) across the two views as ``consistent'' and those with different predicted labels as ``inconsistent''. We note that while invariance across augmented views has been used extensively in prior work, particularly in recent work on semi-~\cite{sajjadi2016regularization,berthelot2019mixmatch,xie2020unsupervised,sohn2020fixmatch} and self-supervised ~\cite{chen2020simple,he2020momentum,xie2020propagate} learning, we instead propose using such predictive consistency to \emph{detect} reliable predictions on which to self-train. In Sec.~\ref{sec:experiments} we empirically demonstrate that such consistency is indeed a reliable proxy for correctness.

\noindent\textbf{ Class-conditioned confidence thresholding.} 
In addition to predictive consistency, we also aim to capture a notion of the intrinsic model confidence. Ideally, the instances on which the model is highly confident should be trusted as reliable. However, a single static threshold will be inadequate for this goal since output probability distributions differ by category~\cite{zou2019confidence}. Therefore, we compute a per-category empirical range to choose an adaptive per-category confidence threshold. Given a batch, we gather all output probabilities and select a confidence threshold per category, $t_c \in \mathbb{R}$, corresponding to the top K-th percentile (K=50 in our experiments) of observed confidence values for category $c$. We consider an output prediction to be high confidence if its top score is greater than the corresonding category threshold: $\max \outputProb_{ij} > t_{\argmax \outputProb_{ij}}$.  
Confidence thresholding is performed over the second view, $\Vb$, as we later opt to self-train on this view. 

Overall, for a pixel, $\imgInput_{ij}$, with per-view probabilistic and categorical predictions, $\outputProb, \Va$ and $\tilde{\outputProb},\Vb$, we define a binary reliability value, $r_{ij}$, in the following way:
\begin{equation}
	r_{ij} = 
	\begin{cases}
		1 & \text{if } \overbrace{\Va_{ij} = \Vb_{ij}}^{\text{consistent}} \; 
		\text{or} \; 
		\overbrace{\max\; \tilde{\outputProb}_{ij} > t_{\Vb_{ij}}}^{\text{confident}} \\
      0  & \text{otherwise} \\
	\end{cases}
\end{equation}

\subsubsection{Learning Objectives}

Using the reliability measure defined above, we now specify a set of learning objectives for each unlabeled target pixel. For pixels deemed reliable, we make the assumption (and empirically confirm -- Sec.~\ref{sec:analysis}) that the current pixel pseudolabel is likely to be accurate. Thus, we train using a standard self-training cross-entropy loss on the predicted pseudolabel. Notably this is different from prior approaches which optimize pseudolabel predictions over all samples~\cite{vu2019advent,chen2019domain,li2020rethinking,wang2021tent} or on those with high model confidence alone~\cite{zou2018unsupervised,tan2019generalized}. Meanwhile, pixels deemed unreliable are excluded from self-training.

\noindent \textbf{Selective self-training.} Having obtained pseudolabels and reliability assignments, we update model parameters via self-training. In a source-free setting, we do not have access to any source data or labels and instead only receive a trained source model and unlabeled target data. In such a setting, optimizing all model parameters causes the model to rapidly diverge from its original task. To address this, we update only the model's batch-norm parameters (affine and normalization), as proposed in Wang~\emph{et al.}~\cite{wang2021tent}. To build in data augmentation we opt to backpropagate predictions made on the secondary view ($\Vb$).

Further, to address the significant label imbalance across categories prevalent in semantic segmentation datasets~\cite{zou2018unsupervised}, we maintain a running average of model predictions over the last-$Q$ batches, denoted by $q \in \mathbb{R}^{\numCls}$ (for $\numCls$ categories), and employ log-inverse frequency loss-weighting: We compute a per category loss weight $\lambda_c$ based on its normalized log inverse frequency~\cite{ramos2003using}: 
$\lambda_c = \log \left[\frac{\Sigma_{c=1}^{\numCls}q_c}{q_c^{\eta}} \right]$, 
where $\eta<1$ is a damping factor. We then minimize a cross-entropy loss $L_{CE}$ over reliable predictions. The self-training objective we minimize is:

\begin{equation}
	L_{SST}(\imgInput_{ij}) =  r_{ij} \lambda_{\Vb_{ij}} L_{CE}(\tilde{\outputProb}_{ij}, \Vb_{ij})
\end{equation}

Finally, to encourage the model to make diverse predictions over the target domain, we add a target ``information entropy'' loss ${L}_{IE}$ proposed in Li ~\etal~\cite{li2020rethinking}, which helps prevent trivial solutions when self-training in the presence of label imbalance. To implement this, we update the model to maximize entropy over the running average of its predictions $q$. ${L}_{IE}$ is given by: $L_{IE}(\imgInput_{ij})  = \sum_{c=1}^{\numCls} \tilde{\outputProb}_{ijc} \log q_c$

For ${L}_{IE}$ loss weight $\alpha$, the complete \method loss objective that is backpropagated is given by:
\begin{equation}
    \scalebox{0.875}[1]{$L_{\method} = \mathbb{E}_{\imgInput \sim \mathcal{P}_\target}\left[ \frac{1}{HW} \sum_{i=1,j=1}^{H,W} L_{SST}(\imgInput_{ij}) + \alpha L_{IE}(\imgInput_{ij})\right]$}
\end{equation}    

\noindent\textbf{Connection to theory.} Wei~\emph{et al.}~\cite{wei2020theoretical} recently proposed a theoretical framework to understand the effectiveness of self-training, 
under two data assumptions: i) \emph{expansion}, meaning that for each ground truth class, a low probability subset of its data must contain a neighborhood atleast $c$ times as large ($c$ is the expansion factor) that it can expand to, with neighborhoods defined based on data augmentations, and ii) \emph{separability}, stating that neighborhoods in different classes should have minimal overlap. Under these assumptions, they show that self-training in combination with input consistency regularization can \emph{denoise} the pseudolabeler. Formally, consider source model $\model_\source$ with initial target error $err_{\target}(h_\source)$ and final model $\model$; it can be shown that $err_{\target}(\model) \leq \frac{2}{c-1} \cdot err_{\target}(\model_\source)$ (Theorem 4.3~\cite{wei2020theoretical}).

For UDA, they assume $c=3$, and an initial pseudolabeler error rate $<\frac{1}{3}$ for each class. However, under a severe domain shift, it is easy to show that the intial source model may have \emph{significantly} higher per-class error rates \emph{e.g.} for GTA$\to$Cityscapes, 8/19 categories have a pixel accuracy lower than 33\%. Our approach circumvents this challenge by optimizing only on \emph{select} augmented pseudolabels that have a high likelihood of being correct.
\vspace{-5pt}
\section{Experiments}
\label{sec:experiments}
\vspace{-3pt}

\subsection{Setup}

\noindent We evaluate \method on 3 shifts for segmentation adaptation.

\noindent \textbf{GTA5$\to$Cityscapes.} 
GTA5~\cite{richter2016playing} contains $\sim$25k synthetic images (of resolution 1914x1052) with pixel-level annotations extracted from the GTA5 game. Cityscapes~\cite{cordts2016cityscapes} comprised of real images (of resolution 2048x1024) of daytime street scenes from 50 European cities. Following prior work~\cite{hoffman2017cycada,vu2019advent,zou2019confidence}, we use the unlabeled train split of Cityscapes for DA (2975 images) and report 19-way classification performance over its validation split (500 images). 

\noindent \textbf{SYNTHIA$\to$Cityscapes.} SYNTHIA~\cite{ros2016synthia} is a semantic segmentation dataset comprised of 9400 synthetic scenes of size 1280x760. 16 categories overlap between SYNTHIA and Cityscapes; following prior work we report performance after adaptation for 16-way and 13-way classification (excluding 3 challenging categories: wall, fence, and pole).

\noindent \textbf{Cityscapes$\to$Dark Zurich Night.} The Dark Zurich dataset~\cite{sakaridis2019guided} consists of 2416 nighttime, 2920 twilight, and 3041 daytime images of resolution 1920x1080. We only use the 2416 nighttime images for adaptation (skipping twilight and daytime), tune performance on the Dark-Zurich validation set (50 images), and report adaptation performance on the Dark-Zurich test set (151 images).  

\noindent \textbf{Metric.} We report per-category Intersection-over-Union (IoU) and its mean across classes (mIoU).

\begin{table}[b]
    \begin{center}
    \resizebox{\linewidth}{!}{
    \begin{tabular}{lcccc}
    \toprule
    \textbf{Method} & \textbf{optim params} & \textbf{extra params} &  \textbf{\#losses} & \textbf{DA epochs}  \\
    \midrule    
    SFDA~\cite{liu2021sourcefree} & all & source copy + generator & 5  & 120 \\
    URMA~\cite{fleuret2021uncertainty} & all & aux. decoders & 3 & 6 \\
    \method & BN & -  & 2 & 1 \\
    \bottomrule
    \end{tabular}
    }
    \vspace{-15pt}
    \caption{Conceptual comparison to source-free baselines. BN=batch-norm.}
    \label{tab:baselines}
    \end{center}
\end{table}
\noindent \textbf{Implementation details.} For \method, we only update batch norm (BN) parameters during training, report performance after 1 epoch of adaptation for all settings, and use batch statistics to compute BN normalization parameters at test-time. 
All models are trained with PyTorch~\cite{paszke2019pytorch} with Adam~\cite{kingma2014adam}, with loss weights $\alpha=0.1$, and $\eta=0.5$. 

For GTA5$\to$Cityscapes \& SYNTHIA$\to$Cityscapes, we resize images to 1024x512 on Cityscapes, 1024x560 on GTA5 and 1280x760 on SYNTHIA. Matching Liu~\etal~\cite{liu2021sourcefree}, we use DeepLabV3~\cite{chen2017rethinking} with a ResNet50~\cite{he2016deep} backbone and initialize models with ImageNet~\cite{russakovsky2015imagenet} weights. We train on the source domain for 20 epochs (GTA5) and 10 epochs (SYNTHIA), making use of Gaussian blur and random flip augmentations. We use a batch size of 8, and learning rate of $1\times10^{-4}$ for GTA5 and $5\times10^{-5}$ for SYNTHIA, with weight decay of $5\times10^{-4}$. On SYNTHIA, we L$_2$-normalize per-category classifier weights in the last FC layer and set bias to 0 following~\cite{kang2019decoupling}. See supp. for training details with DeepLabV2. 
For the Cityscapes $\to$ Dark Zurich Night shift, we match Sakaridis ~\etal~\cite{sakaridis2019guided} and use a DeepLabV2~\cite{chen2017deeplab} architecture with ResNet-101~\cite{he2016deep} backbone. We set learning rate to $2.5\times10^{-4}$ and batch size to 8.

\begin{table*}[t]    
    \setlength{\tabcolsep}{3pt}
    \begin{center}
    \resizebox{\columnwidth}{!}{
    
    \begin{tabular}{lcccccccccccccccccccccc}
    \toprule
    \textbf{Method} & \textbf{SF} & \textbf{Arch} & \rotb{road} & \rotb{building} & \rotb{vegetation} & \rotb{car} & \rotb{sidewalk} & \rotb{sky} & \rotb{pole} & \rotb{person} & \rotb{terrain} & \rotb{fence} & \rotb{wall} & \rotb{bicycle} & \rotb{sign} & \rotb{bus} & \rotb{truck} & \rotb{rider} & \rotb{light} & \rotb{train} & \rotb{motorcycle} &  \textbf{mIoU}  \\
    \midrule
    source & - & A & 74.6	&77.6&	77.9&	71.1&	16.0&	79.4 &	14.8&	57.9&	9.0&	16.6&	12.5&	11.0 &	15.1&	18.8&	25.0 &	24.1 &	35.6 &	0.5&	15.4&	34.4 \\
    Test-time BN~\cite{nado2020evaluating} & \ding{51} &A & 79.5 & 79.5 & 81.3 & 72.7 & 29.8 & 74.1 & 28.0 & 58.8 & 25.3 & 22.2 & 19.4 & 11.0& 22.6 & 17.0 & 24.5 & 19.9 & 34.3 & 2.4 & 14.7 & 37.7 \\
    \texttt{TENT}~\cite{wang2021tent} & \ding{51}& A& 87.3 &     79.8 &       83.8 &  85.0 &     39.0 &  77.7 &  21.2 &   57.9 &    34.7 &  19.6 &  24.3 &     4.5 &  16.6 &  20.8 &  24.9 &  17.8 &  25.1 &   2.0 &      16.6 &  38.9  \\
    \texttt{SFDA}~\cite{liu2021sourcefree} & \ding{51}&A&84.2	& 82.7	& 82.4	& 80.0	& 39.2	& 85.3	& 25.9	& 58.7	& 30.5	& 22.1	& 27.5	& 30.6	& 21.9	& 31.5 & 33.1	& 22.1	& 31.1	& 3.6 & 27.8 & 43.2  \\
    \rowcolor{Gray}
    \method (ours) & \ding{51} & A & 92.6	& 84.6&	86.8 &	84.8 &	56.1 &	82.2 &	45.3 & 63.4& 43.8	& 28.7	&	26.8 &	14.0 &41.1	&34.9& 16.7 &	29.6 &45.7 &5.6	&12.6&	\textbf{47.1} \\
    \midrule
    source & - & B & 69.7	& 73.3&	78.7 &	70.6 &20.5	&	68.2 &	23.5 & 53.9 &	18.7&	12.3 &	22.1 & 31.5	&	17.9 & 4.5 & 32.2 &	26.5 & 31.8 &	8.1 & 26.8 &	36.4 \\
    \texttt{URMA~\cite{fleuret2021uncertainty}} & \ding{51} & B & 92.3	& 81.6 &	84.2 &	81.7 & 55.2	&83.8	 & 37.1	& 57.7 & 35.9	& 18.8	&	30.8& 40.4	& 12.1	&  44.3 & 27.5 &	24.1 & 17.7 &	6.9& 24.1 &	45.1 \\
    \rowcolor{Gray}
    \method \textbf{(ours)} & \ding{51} &  B & 90.3	& 81.8 &	83.6 &	84.7 &	41.2 &	79.7 & 34.5	& 61.4 & 34.6 	& 21.4	&26.5	&34.6	&33.3	&39.5 & 30.3 &	19.3 &40.4 &	7.3 & 27.6 &	\textbf{45.9} \\
    \midrule
    \midrule
    \texttt{MaxSquares~\cite{chen2019domain}} & \ding{55} &A& 85.8&	82.4&	83.2&	81.0&	33.6&	79.8&	26.5 &	57.8&	32.9&	25.0 &	25.3&	32.4 &	18.7&	32.6 &	32.1&	22.2&	33.3&	5.2&	29.8 & 43.1 \\
    \texttt{IAST} & \ding{55} &B &  94.1	& 85.4&	84.8 &87.6	 &58.8	&	88.7 &25.1	& 62.7&34.6	&29.2	&39.7	&40.2	&34.2	&50.3& 42.3& 30.3 &43.1 &24.7	& 35.2&	52.2 \\
    \bottomrule
    \end{tabular}}
    \vspace{-15pt}
    \caption{\textbf{GTA5$\to$Cityscapes}: IoU on the Cityscapes validation set. Categories are in descending order of frequency. \textbf{SF} = Source Free. A=DeepLabV3 with ResNet50 backbone. B=DeepLabV2 with ResNet101 backbone. 
    }
    \label{tab:results_g2c}
    \end{center}
\end{table*}

\begin{table*}[t]
    \setlength{\tabcolsep}{3pt}
    \begin{center}
    \resizebox{\columnwidth}{!}{
    \begin{tabular}{lcccccccccccccccccccc}
    \toprule
    \textbf{Method} & \textbf{SF} & \textbf{Arch} & \rotb{road} & \rotb{building} & \rotb{vegetation} & \rotb{car} & \rotb{sidewalk} & \rotb{sky} & \rotb{pole$^{*}$} & \rotb{person} & \rotb{fence$^{*}$} & \rotb{wall$^{*}$} & \rotb{bicycle} & \rotb{sign} & \rotb{bus} & \rotb{rider} & \rotb{light} & \rotb{motorcycle} &  \textbf{mIoU}  & \textbf{mIoU$^{*}$} \\
    \midrule    
    source & - &A&  50.4&  78.3& 71.8& 44.0& 18.8& 78.4& 24.0& 51.9& 0.2& 2.5& 18.1& 6.5& 5.9& 13.5& 5.3& 0.2& 29.4 & 34.1  \\
    Test-time BN~\cite{nado2020evaluating} & \ding{51} &A& 76.3 & 75.4 & 74.5 & 65.9 & 33.9 & 81.0 & 23.2 & 44.1 & 0.5 & 6.3 & 34.4 & 12.1 & 4.9 & 15.9 & 6.5 & 5.9 & 35.0 & 40.8 \\
    \texttt{TENT}~\cite{wang2021tent} & \ding{51}&A&  88.1 & 74.4 & 77.3 & 77.6 & 44.9 & 82.8 & 21.8 & 52.9 & 0.1 & 4.3 & 15.8 & 7.8 & 7.5 & 9.7 & 2.0 & 0.2 & 35.5 & 41.6  \\    
    \texttt{SFDA}~\cite{liu2021sourcefree}&A& \ding{51} & 81.5& 	80.6& 	83.1& 	81.3& 	43.5& 	87.6& 	19.9& 	36.8& 	0.7& 	1.4& 	31.7& 	7.1& 	22.7& 	9.5& 	4.2& 	8.6& 	39.2 & 	\textbf{45.9} \\
    \rowcolor{Gray}
    \method \textbf{(ours)} &A& \ding{51} &  87.2	& 80.1 &	80.8 &	77.9 &	44.5 &	82.4 &	26.5 & 54.8 &	0.3 & 8.2 & 34.6	&12.6	 & 8.7 &18.8	& 8.2 & 6.0&	\textbf{39.5} & \textbf{45.9} \\
    \midrule
    source & \ding{51} &  B & 45.2	& 72.0 &	75.3 & 39.0	 & 19.6	&81.9	 & 25.4	& 57.3 & 0.1 &	6.7 & 6.7	& 7.8	&	19.5 & 17.3 & 5.5	& 7.0&	31.6 & 36.4 \\
    \texttt{URMA}~\cite{fleuret2021uncertainty} & \ding{51} &  B & 59.3 	&  77.0 &	83.1 &	 76.7 &  24.6	&	80.4 &31.5	& 46.3&	  1.8 &	14.0 & 34.6	&32.0	 & 17.0 & 17.8	& 18.3 & 18.5&	\textbf{39.6} & 45.0 \\
    \rowcolor{Gray}
    \method \textbf{(ours)} & \ding{51} &  B & 74.8	& 79.2 &	78.7 &	74.3 &	32.1 &	83.1 & 29.4	& 57.5 &	0.1 & 5.0 &	39.3& 11.1	 & 20.5 & 26.4	& 3.0 & 12.1 &	39.2 & \textbf{45.5} \\
    \midrule
    \midrule
    \texttt{MaxSquares~\cite{chen2019domain}}  & \ding{55}&A& 81.0 &	82.6&	85.3&	84.7&	39.8&	90.1&	23.2&	39.9&	0.5&	8.7&	33.4&	12.4&	19.4&	8.4&	6.6&	10.2&	39.1 &	45.7 \\
    \texttt{IAST}~\cite{mei2020instance} & \ding{55} &  B & 81.9	& 83.3 &83.4	 &	86.5 &	41.5&85.0	 &32.3	&65.5&4.6	&17.7	&52.7	&28.8	 &38.2 & 30.8	&30.9&33.1 &	49.8  & 57.0 \\
    \bottomrule
    \end{tabular}}
    \vspace{-15pt}
    \caption{\textbf{SYNTHIA$\to$Cityscapes}: IoU on the Cityscapes validation set.  Categories are in descending order of frequency. \textbf{SF} = Source Free. A=DeepLabV3 with ResNet50. B=DeepLabV2 with ResNet101  backbone.  mIoU and mIoU$^{*}$ are calculated over 16 and 13 categories (excluding categories with a $^{*}$).
    }
    \label{tab:results_s2c}
    \end{center}
\end{table*}

\noindent \textbf{Baselines.} \TODO{We use DeepLabV3 with a ResNet50 backbone and compare against}: i) \textbf{TENT}~\cite{wang2021tent}: \texttt{TENT} is a source-free, test-time adaptation algorithm that learns batch-norm parameters so as to minimize predictive entropy over unlabeled target data, ii) \textbf{SFDA}~\cite{liu2021sourcefree}: SFDA is a state-of-the-art source-free DA method that learns an additional generator to simulate source features, followed by a distillation step via a dual-attention module to enable knowledge transfer. This is followed by adaptation via self-training and patch-level self-supervised learning in the target domain. iii) \textbf{Test-time BN}~\cite{nado2020evaluating}, a test-time adaptation method that only updates batch-norm normalization statistics over the entire test dataset before making a prediction. \TODO{We additionally benchmark \method with a DeepLabV2 architecture (ResNet101 backbone) and compare against:} iv) URMA~\cite{fleuret2021uncertainty}, which combines entropy minimization and confidence thresholding-based self-training on pseudolabels, with a robustness objectiveness enforcing stability under feature noising via dropout-equipped auxiliary decoders.

In Table~\ref{tab:baselines} we highlight key differences between \method and our two strongest baselines, SFDA and URMA. Unlike prior work, \method only updates batch-norm parameters, does not require learning additional generators or decoders, has fewer learning objectives making it easier to tune, and converges within a single pass over target data.

\begin{table*}[t]
    \setlength{\tabcolsep}{3pt}
    \begin{center}
    \resizebox{\columnwidth}{!}{
    \begin{tabular}{lccccccccccccccccccccc}
    \toprule
    \textbf{Method} & \textbf{SF} & \rotb{road} & \rotb{building} & \rotb{vegetation} & \rotb{car} & \rotb{sidewalk} & \rotb{sky} & \rotb{pole} & \rotb{person} & \rotb{terrain} & \rotb{fence} & \rotb{wall} & \rotb{bicycle} & \rotb{sign} & \rotb{bus} & \rotb{truck} & \rotb{rider} & \rotb{light} & \rotb{train} & \rotb{motorcycle} &  \textbf{mIoU}  \\
    \midrule
    source & - & 79.0 & 53.0 &43.5&	64.1&	21.8&	18.0	&22.5&	37.4&	10.4&	11.2&	13.3&	7.4&	22.1&	0.0	&6.4&	33.8&	20.2&	52.3&	30.4&	28.8 \\    
    \texttt{Test-Time BN}~\cite{nado2020evaluating} & \ding{51} & 77.9 & 50.0 & 44.6 & 48.4 & 36.8 & 0.6 & 34.8 & 33.8 & 7.0 & 15.0 & 19.8 & 15.4 & 25.9 & 0.0 & 0.1 & 38.1 & 16.9 & 36.0 & 30.9 & 28.0 \\    
    \texttt{TENT}~\cite{wang2021tent} & \ding{51} & 79.1& 43.8& 13.5& 61.0& 31.9& 0.0& 29.1& 39.3& 0.8& 14.1& 8.7& 12.8& 23.2& 0.0& 19.1& 30.2& 17.1& 51.5& 30.7 & 26.6 \\    
    \rowcolor{Gray}
    \method \textbf{(ours)} & \ding{51} & 85.4 & 48.6 & 57.8 & 61.3 & 47.2 & 0.4 & 34.6 & 34.5 & 14.0 & 16.9 & 30.7 & 18.3 & 27.8 & 0.0 & 7.1 & 30.9 & 16.5 & 53.7 & 29.7 & \textbf{32.4} \\    
    \midrule
    \midrule
    \texttt{AdaptSegNet~\cite{tsai2018learning}} & \ding{55} & 86.1	& 55.1& 	37.2& 	68.2& 	44.2& 	1.2& 	21.1& 	35.9& 	8.4& 	4.8& 	22.2& 	15.6& 	16.7& 	0.0 & 	45.1& 	26.7& 	5.6& 	50.1& 	33.9& 	30.4 \\
    \texttt{Advent~\cite{vu2019advent}} & \ding{55} & 85.8	&55.5&	32.1&	64	&37.9&	2&	23.1&	39.9&	8.7&	14.5&	27.7&	20.7&	21.1&	0.0 &	13.8&	16.6&	14&	58.8&	28.5&	29.7 \\
    \bottomrule
    \end{tabular}}
    \vspace{-15pt}
    \caption{\textbf{Cityscapes$\to$Dark Zurich Night}: IoU on the DZ-Night test set. Categories in descending order of frequency. \textbf{SF} = Source Free.
    }
    \label{tab:results_c2d}
    \end{center}
\end{table*}

For context and completeness, we also report performance for a state-of-the-art source-\emph{equipped} baseline for each architecture: i) For DeepLabV3, we copy results for \textbf{MaxSquares}~\cite{chen2019domain} from Liu ~\emph{et al.}~\cite{liu2021sourcefree}. ii) For DeepLabV2, we include results for IAST~\cite{mei2020instance}. These baselines prevent task drift by minimizing an additional supervised loss on labeled source data.

\noindent \textbf{GTA5$\to$Cityscapes.} Table~\ref{tab:results_g2c} presents our results for adaptation from GTA5$\to$Cityscapes. We evaluate our method (\method) after a \emph{single pass} over the unlabeled target data (\emph{i.e.} one epoch) and use multi-scale evaluation following prior work~\cite{mei2020instance}. With DeepLabV3, \method outperforms the state-of-the-art SFDA~\cite{liu2021sourcefree} by 3.9 points, and on 13/19 categories. \method also significantly outperforms \texttt{TENT}~\cite{wang2021tent} (+8.2). With DeepLabV2, \method outperforms the state-of-the-art URMA~\cite{fleuret2021uncertainty} method by 0.8 mIoU without training auxiliary decoders (Table~\ref{tab:baselines}). Overall, we observe lower performance with DeepLabV2 despite a deeper backbone and better source performance due to DeepLabV3's superior learned upsampling strategy, which complements our pixel-level predictive consistency scheme.
We note that source-equipped methods like IAST~\cite{mei2020instance} achieve significantly higher performance by leveraging labeled source data; we thus restrict our state-of-the-art performance claims strictly to the source-free setting.

\noindent\textbf{SYNTHIA$\to$Cityscapes.} Table~\ref{tab:results_s2c} presents our results for adaptation from SYNTHIA$\to$Cityscapes. With DeepLabV3, \method achieves an mIoU/mIoU* of 39.5/45.9, narrowly outperforming SFDA (39.2/45.9). We note here that SFDA makes use of 120 epochs of adversarial learning whereas \method is only trained for a single epoch of adaptation. Further, \method also considerably outperforms \texttt{TENT} (by 4.0/4.3 points). 
\TODO{With DeepLabV2, \method obtains 39.2/45.5 mIoU/mIoU*,  matching URMA (39.6/45.0). We note that this is without learning auxiliary decoders, and despite a much lower source performance (31.6/36.4), as we were unable to reproduce URMA's source performance (34.9/40.3).}

\noindent\textbf{Cityscapes$\to$Dark Zurich Night.} See Table~\ref{tab:results_c2d}. \method again improves upon \texttt{TENT}~\cite{wang2021tent} (+5.8) and \texttt{Test-time BN}~\cite{nado2020evaluating} (+4.4 ). Note that we restrict comparisons to methods that only use nighttime images for DA.

\subsection{Ablating \method}
\vspace{-5pt}

In Table~\ref{tab:ablations} we present ablations of \method for both architectures. We observe:

\begin{table}[!hbtp]
    \resizebox{\linewidth}{!}{
    \setlength{\tabcolsep}{2pt}
    \begin{tabular}{cccccccc}
        \toprule
         \multicolumn{1}{c}{\centering \bf Reg.}&  \multicolumn{2}{c}{\centering \bf Selection strategy} & \multicolumn{1}{c}{\centering \bf Class bal.} & \multicolumn{2}{c}{\centering \bf Loss} & \multicolumn{2}{c}{\centering \bf mIoU $\uparrow$} \\
        \cmidrule(l{4pt}r{4pt}){1-1}
        \cmidrule(l{4pt}r{4pt}){2-3}
        \cmidrule(l{4pt}r{4pt}){4-4}
        \cmidrule(l{4pt}r{4pt}){5-6}
        \cmidrule(l{4pt}r{4pt}){7-8}
         {\centering H(Y) loss }&  {\centering Confidence }& {\centering Consistency} & {\centering Loss wts.} & {\centering Reliable }& {\centering Unreliable} & {\centering DLV3} & {\centering DLV2} \\ 
        \toprule                 
        \multicolumn{4}{c}{\multirow{1}{*}{  \centering (Unconstrained self-training on all predictions)}} & CE & CE & 34.04 & 32.28 \\
        \midrule         
        \ding{51} &  &  &  & CE & CE & 39.11 &  38.75 \\
        \ding{51} & \ding{51} &  &  & CE & None &  16.93 & 24.91 \\           
        \ding{51} &  & \ding{51} &  & CE & None & 47.05 & 44.94 \\           
        \ding{51} & \ding{51} & \ding{51} &  & CE & None & 47.12  & 45.43 \\                        
        \midrule
        \rowcolor{Gray}
         \ding{51} & \ding{51} & \ding{51} &\ding{51}  &    CE & None &  47.12 & 45.89 \\                
        \bottomrule
        \end{tabular}}
        \vspace{-2pt}
        \caption{Ablating \method on GTA$\to$Cityscapes. We report mIoU over all categories for the DeepLabV3 w/ ResNet50 and DeepLabV2 w/ ResNet101 architectures. CE  = cross-entropy against predicted pseudolabel.}
        \vspace{-5pt}
      \label{tab:ablations}
\end{table}

\noindent\textbf{$\triangleright$ Unconstrained self-training leads to suboptimal performance (Row 1).} We first try self-training on all pixels by minimizing cross-entropy with respect to predictions. As seen, this achieves 34.04/32.28 mIoU, underperforming even the source model (mIoU=34.4/36.4, Tab.~\ref{tab:results_g2c}).

\noindent\textbf{$\triangleright$ Pixel-level predictive consistency is an effective selection strategy (Row 3-6).} To regularize self-training, we first add a target information entropy regularizer (Sec~\ref{sec:approach}) and find that   mIoU increases to 39.11/38.75 (Row 2). Next, to validate our selection criterion, we first use only confidence for selection (we select predictions in the top-50 \%ile by confidence per-category) -- despite careful tuning this leads to a low mIoU of 16.93/24.91 (Row 3), indicating that confidence alone is a poor indicator of reliability. 

Next, we try using predictive consistency for selection, and find this improves mIoU to 47.05/44.94 (Row 4), validating the hypothesis that predictive consistency is an effective proxy for reliability (more analysis in Sec.~\ref{sec:analysis}). Finally, combining consistency and confidence obtains the same performance on DLV3 but improves DLV2 mIoU to 45.43 (Row 5), which loss weighting further improves to 45.89.  In the supplementary, we also try other combinations of consistency and confidence. %

\begin{figure*}[t]
    \centering
    \includegraphics[width=\linewidth]{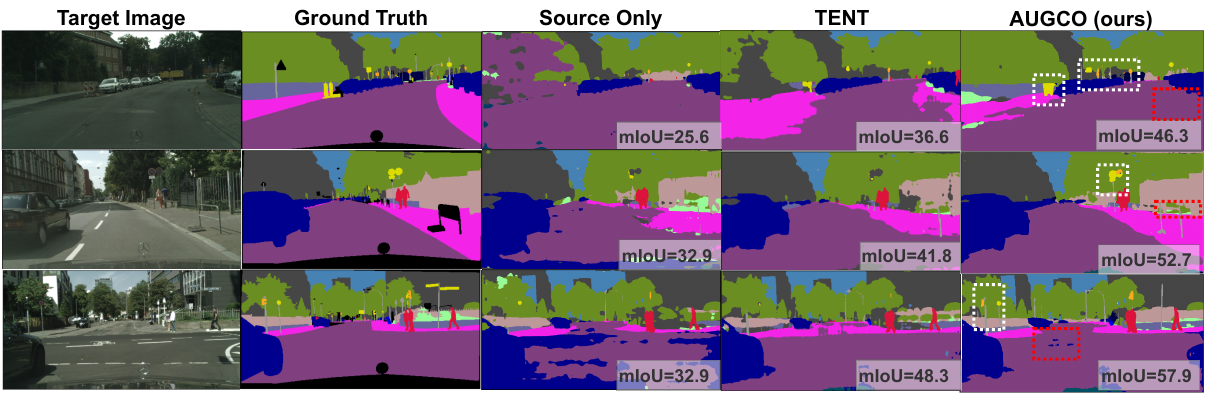}
    \caption{Qualitative segmentation results of the source model, \texttt{TENT}~\cite{wang2021tent}, and \method. White boxes highlight categories recovered by \method, whereas red boxes show some failure cases.}
    \label{fig:qual}
    \vspace*{-7pt}
 \end{figure*}
 
\noindent\textbf{$\triangleright$ Upper bound: Reliability oracle.} As an upper bound, we train our method with perfect reliability maps (using ground truth labels) and find performance converges to an mIoU of 50.79 within two epochs -- the 3.7 mIoU gain over our method captures the performance drop due to imperfect reliability maps. In supplementary we analyze the relationship between reliabilty mapping accuracy and mIoU.

\noindent\textbf{$\triangleright$ Varying optimization parameters.} We now try alternatively training all model parameters instead of just batch norm. We observe rapid task with larger learning rates and carefully tune optimizers  to obtain an mIoU of 46.74 with a learning rate of 5$\times10^{-6}$ and weight decay of 5$\times10^{-4}$, worse than when training batch-norm parameters alone. In supp. we also try training all parameters with a source-equipped version of our method.

\noindent\textbf{Ablating augmentations.} In Table~\ref{tab:aug_ablations} we specifically ablate the choice of augmentations used for measuring predictive consistency and report mIoU with DeepLabV3. We observe:

\begin{table}[b]
    \resizebox{\linewidth}{!}{
    \begin{tabular}{cccc}
        \toprule
        \multicolumn{1}{c}{\centering \bf transforms} &  \multicolumn{1}{c}{\centering \bf use to train} & \multicolumn{1}{c}{\centering \bf use to select} & \textbf{mIoU} \\        
        \toprule                 
        none & & & 29.7 \\
        \midrule
        spatial & \ding{51} & & 31.6 \\
        spatial & \ding{51} & \ding{51} & 44.6 \\
        \midrule
        pixel & \ding{51} &  & 30.2 \\
        pixel & \ding{51} & \ding{51} & 45.6 \\        
        \midrule
        spatial+pixel & \ding{51} &  & 33.2 \\                
        \rowcolor{Gray}
        spatial+pixel & \ding{51} & \ding{51} & 47.1 \\        
        \bottomrule
        \end{tabular}}
        \caption{Augmentation ablations. Gray row is our method.}
        \label{tab:aug_ablations}
\end{table}

\noindent\textbf{$\triangleright$ Gains are not just from data augmentation.} We vary the set of transforms across spatial (crop and resize), pixel (color jitter), or both. Across all settings, we find that using predictive consistency under each transform to also \emph{select} pixels is far more important than just using the transforms for training itself, \emph{e.g.} +13.9 points better when using both spatial and pixel transforms (last 2 rows).

\noindent\textbf{$\triangleright$ Spatial and pixel transforms are complementary.} While consistency under both spatial and pixel transformations is effective for selection (44.6 and 45.6 points), using both simultaneously performs best (47.1 mIoU).

\subsection{Analyzing \method}
\label{sec:analysis}
\vspace{-5pt}

Fig.~\ref{fig:qual} shows qualitative segmentation results achieved by our method. We further analyze: 

\begin{figure}[t]
    \centering    
      \begin{subfigure}[b]{0.47\textwidth}
        \centering
        \includegraphics[width=.95\textwidth]{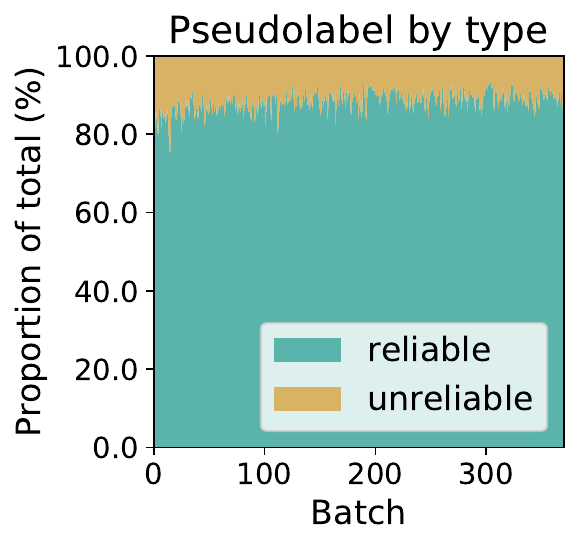}
        \caption[]%
        {{\small \% pseudolabel types vs iter.}}  
        \label{fig:breakdown}
        \end{subfigure}
        \begin{subfigure}[b]{0.47\textwidth}
            \centering
            \includegraphics[width=1.0\linewidth]{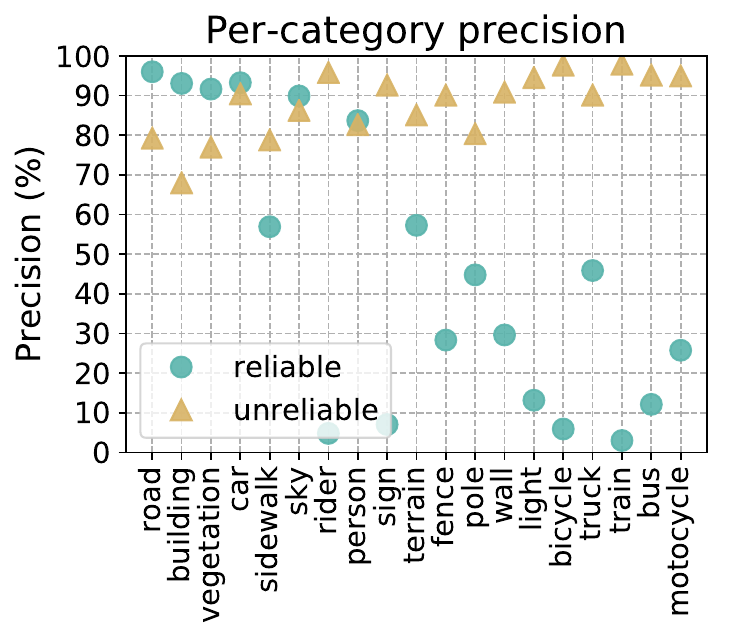}
            \caption[]%
            {{\small Reliability Precision }}
            \label{fig:precision}
            \end{subfigure}
            \vspace{-7pt}
      \caption{Analyzing pseudolabel types and reliability precision.}
      \label{fig:analysis}
  \end{figure}
 
\noindent \textbf{Evaluating reliability measure (Fig.~\ref{fig:analysis}).} To evaluate whether \method's selection strategy is indeed a good indicator of reliability, we first measure the accuracy of pseudolabels marked as reliable and unreliable -- reliable pseudolabels have an accuracy of \textbf{86.2}\%, whereas unreliable ones have a low accuracy of \textbf{19.1}\%; further these statistics are stable over the course of training (plot in supp.). Fig.~\ref{fig:breakdown} breaks down predictions as reliable and unreliable as training progresses -- as seen, the proportion of reliable pixels increases over time.

In Fig.~\ref{fig:precision}, we evaluate the reliability measure by category. For each category, we report i) \emph{precision} of reliability with respect to correctness (when a pixel prediction is reliable, how often is it actually correctly classified?), and ii) precision of unreliability with respect to incorrectness. As seen, unreliable predictions are highly correlated with being incorrect across categories, which explains the effectiveness of excluding them from training. However, the precision of the reliability measure is significantly higher for head categories (\emph{e.g.} road, building, car) than for the tail (\emph{e.g.} bicycle, bus, train). In fact, we observe a high correlation between the precision of reliabilty for a given category and its mIoU after adaptation (Pearson correlation coefficient=0.88).

\noindent\textbf{Computational efficiency.} \method is a lightweight adaptation method, requiring i) only one additional forward pass for the secondary view during training, and ii) updates only to batchnorm parameters. 
Further, across our experiments, we observe \method to lead to optimal performance within a single pass over target data.

\noindent\textbf{Convergence.} To test \method's training stability beyond an epoch, we perform GTA$\to$Cityscapes adaptation for 8 epochs with a lower learning rate of 1$\times 10^{-5}$. We observe performance peaks at 45.48 at 5 epochs and then remains stable for the remaining epochs.

\section{Limitations \& Conclusion}
\label{sec:conclusion}

Despite \method's versatility, its effectiveness is limited by the correlation between predictive consistency and correctness, which varies across categories. Modifying source training to improve this correlation may be promising future work. Further, \method requires an extra forward pass and confidence thresholding, which makes it slightly slower than \texttt{TENT}~\cite{wang2021tent}. Finally, while \method seeks to make source-free adaptation more robust, such unsupervised training still has the potential to fail silently. A deeper investigation into the the failure modes of such methods is required before deployment in sensitive applications. Despite these limitations, we believe that \method is a step towards developing an effective source-free adaptation algorithm that can be used in the real-world at practical computational cost.

\noindent\textbf{Acknowledgements.} We thank Prithvijit Chattopadhyay for feedback on the draft. This work was supported in part by funding from the DARPA LwLL project.

{\small
\bibliographystyle{ieeetr}
\bibliography{main}

\begin{thebibliography}{10}

\bibitem{liang2020we}
J.~Liang, D.~Hu, and J.~Feng, ``Do we really need to access the source data?
  source hypothesis transfer for unsupervised domain adaptation,'' in {\em
  International Conference on Machine Learning}, pp.~6028--6039, PMLR, 2020.

\bibitem{li2020model}
R.~Li, Q.~Jiao, W.~Cao, H.-S. Wong, and S.~Wu, ``Model adaptation: Unsupervised
  domain adaptation without source data,'' in {\em Proceedings of the IEEE/CVF
  Conference on Computer Vision and Pattern Recognition}, pp.~9641--9650, 2020.

\bibitem{kim2020domain}
Y.~Kim, S.~Hong, D.~Cho, H.~Park, and P.~Panda, ``Domain adaptation without
  source data,'' {\em arXiv preprint arXiv:2007.01524}, 2020.

\bibitem{kundu2020universal}
J.~N. Kundu, N.~Venkat, R.~V. Babu, {\em et~al.}, ``Universal source-free
  domain adaptation,'' in {\em Proceedings of the IEEE/CVF Conference on
  Computer Vision and Pattern Recognition}, pp.~4544--4553, 2020.

\bibitem{wang2021tent}
D.~Wang, E.~Shelhamer, S.~Liu, B.~Olshausen, T.~Darrell, U.~Berkeley, and
  A.~Research, ``Tent: Fully test-time adaptation by entropy minimization,'' in
  {\em International Conference on Learning Representations}, vol.~4, p.~6,
  2021.

\bibitem{liu2021sourcefree}
Y.~Liu, W.~Zhang, and J.~Wang, ``Source-free domain adaptation for semantic
  segmentation,'' in {\em Proceedings of the IEEE conference on computer vision
  and pattern recognition}, 2021.

\bibitem{saenko2010adapting}
K.~Saenko, B.~Kulis, M.~Fritz, and T.~Darrell, ``Adapting visual category
  models to new domains,'' in {\em European conference on computer vision},
  pp.~213--226, Springer, 2010.

\bibitem{ganin2014unsupervised}
Y.~Ganin and V.~Lempitsky, ``Unsupervised domain adaptation by
  backpropagation,'' in {\em International Conference on Machine Learning},
  pp.~1180--1189, 2015.

\bibitem{long2015learning}
M.~Long, Y.~Cao, J.~Wang, and M.~Jordan, ``Learning transferable features with
  deep adaptation networks,'' in {\em International Conference on Machine
  Learning}, pp.~97--105, 2015.

\bibitem{hoffman2017cycada}
J.~Hoffman, E.~Tzeng, T.~Park, J.-Y. Zhu, P.~Isola, K.~Saenko, A.~Efros, and
  T.~Darrell, ``Cycada: Cycle-consistent adversarial domain adaptation,'' in
  {\em International Conference on Machine Learning}, pp.~1989--1998, 2018.

\bibitem{vu2019advent}
T.-H. Vu, H.~Jain, M.~Bucher, M.~Cord, and P.~P{\'e}rez, ``Advent: Adversarial
  entropy minimization for domain adaptation in semantic segmentation,'' in
  {\em Proceedings of the IEEE/CVF Conference on Computer Vision and Pattern
  Recognition}, pp.~2517--2526, 2019.

\bibitem{fleuret2021uncertainty}
F.~Fleuret {\em et~al.}, ``Uncertainty reduction for model adaptation in
  semantic segmentation,'' in {\em Proceedings of the IEEE/CVF Conference on
  Computer Vision and Pattern Recognition}, pp.~9613--9623, 2021.

\bibitem{grandvalet2005semi}
Y.~Grandvalet, Y.~Bengio, {\em et~al.}, ``Semi-supervised learning by entropy
  minimization.,'' in {\em CAP}, pp.~281--296, 2005.

\bibitem{chen2019progressive}
C.~Chen, W.~Xie, W.~Huang, Y.~Rong, X.~Ding, Y.~Huang, T.~Xu, and J.~Huang,
  ``Progressive feature alignment for unsupervised domain adaptation,'' in {\em
  Proceedings of the IEEE/CVF Conference on Computer Vision and Pattern
  Recognition}, pp.~627--636, 2019.

\bibitem{jiang2020implicit}
X.~Jiang, Q.~Lao, S.~Matwin, and M.~Havaei, ``Implicit class-conditioned domain
  alignment for unsupervised domain adaptation,'' in {\em International
  Conference on Machine Learning}, pp.~4816--4827, PMLR, 2020.

\bibitem{prabhu2020sentry}
V.~Prabhu, S.~Khare, D.~Kartik, and J.~Hoffman, ``Sentry: Selective entropy
  optimization via committee consistency for unsupervised domain adaptation,''
  in {\em Proceedings of the IEEE/CVF International Conference on Computer
  Vision}, pp.~8558--8567, 2021.

\bibitem{tan2019generalized}
S.~Tan, X.~Peng, and K.~Saenko, ``Class-imbalanced domain adaptation: An
  empirical odyssey,'' in {\em Proceedings of the European Conference on
  Computer Vision (ECCV) Workshops}, September 2020.

\bibitem{snoek2019can}
J.~Snoek, Y.~Ovadia, E.~Fertig, B.~Lakshminarayanan, S.~Nowozin, D.~Sculley,
  J.~Dillon, J.~Ren, and Z.~Nado, ``Can you trust your model's uncertainty?
  evaluating predictive uncertainty under dataset shift,'' in {\em Advances in
  Neural Information Processing Systems}, pp.~13969--13980, 2019.

\bibitem{chen2020simple}
T.~Chen, S.~Kornblith, M.~Norouzi, and G.~Hinton, ``A simple framework for
  contrastive learning of visual representations,'' in {\em International
  conference on machine learning}, pp.~1597--1607, PMLR, 2020.

\bibitem{he2020momentum}
K.~He, H.~Fan, Y.~Wu, S.~Xie, and R.~Girshick, ``Momentum contrast for
  unsupervised visual representation learning,'' in {\em Proceedings of the
  IEEE/CVF Conference on Computer Vision and Pattern Recognition},
  pp.~9729--9738, 2020.

\bibitem{xie2020propagate}
Z.~Xie, Y.~Lin, Z.~Zhang, Y.~Cao, S.~Lin, and H.~Hu, ``Propagate yourself:
  Exploring pixel-level consistency for unsupervised visual representation
  learning,'' {\em arXiv preprint arXiv:2011.10043}, 2020.

\bibitem{richter2016playing}
S.~R. Richter, V.~Vineet, S.~Roth, and V.~Koltun, ``Playing for data: Ground
  truth from computer games,'' in {\em European conference on computer vision},
  pp.~102--118, Springer, 2016.

\bibitem{cordts2016cityscapes}
M.~Cordts, M.~Omran, S.~Ramos, T.~Rehfeld, M.~Enzweiler, R.~Benenson,
  U.~Franke, S.~Roth, and B.~Schiele, ``The cityscapes dataset for semantic
  urban scene understanding,'' in {\em Proceedings of the IEEE conference on
  computer vision and pattern recognition}, pp.~3213--3223, 2016.

\bibitem{sakaridis2019guided}
C.~Sakaridis, D.~Dai, and L.~V. Gool, ``Guided curriculum model adaptation and
  uncertainty-aware evaluation for semantic nighttime image segmentation,'' in
  {\em Proceedings of the IEEE/CVF International Conference on Computer
  Vision}, pp.~7374--7383, 2019.

\bibitem{ros2016synthia}
G.~Ros, L.~Sellart, J.~Materzynska, D.~Vazquez, and A.~M. Lopez, ``The synthia
  dataset: A large collection of synthetic images for semantic segmentation of
  urban scenes,'' in {\em Proceedings of the IEEE conference on computer vision
  and pattern recognition}, pp.~3234--3243, 2016.

\bibitem{hoffman2016fcns}
J.~Hoffman, D.~Wang, F.~Yu, and T.~Darrell, ``Fcns in the wild: Pixel-level
  adversarial and constraint-based adaptation,'' {\em arXiv preprint
  arXiv:1612.02649}, 2016.

\bibitem{tsai2018learning}
Y.-H. Tsai, W.-C. Hung, S.~Schulter, K.~Sohn, M.-H. Yang, and M.~Chandraker,
  ``Learning to adapt structured output space for semantic segmentation,'' in
  {\em Proceedings of the IEEE conference on computer vision and pattern
  recognition}, pp.~7472--7481, 2018.

\bibitem{luo2019significance}
Y.~Luo, P.~Liu, T.~Guan, J.~Yu, and Y.~Yang, ``Significance-aware information
  bottleneck for domain adaptive semantic segmentation,'' in {\em Proceedings
  of the IEEE/CVF International Conference on Computer Vision}, pp.~6778--6787,
  2019.

\bibitem{murez2018image}
Z.~Murez, S.~Kolouri, D.~Kriegman, R.~Ramamoorthi, and K.~Kim, ``Image to image
  translation for domain adaptation,'' in {\em Proceedings of the IEEE
  Conference on Computer Vision and Pattern Recognition}, pp.~4500--4509, 2018.

\bibitem{hoffman2018cycada}
J.~Hoffman, E.~Tzeng, T.~Park, J.-Y. Zhu, P.~Isola, K.~Saenko, A.~Efros, and
  T.~Darrell, ``Cycada: Cycle-consistent adversarial domain adaptation,'' in
  {\em International conference on machine learning}, pp.~1989--1998, PMLR,
  2018.

\bibitem{kang2020pixel}
G.~Kang, Y.~Wei, Y.~Yang, Y.~Zhuang, and A.~G. Hauptmann, ``Pixel-level cycle
  association: A new perspective for domain adaptive semantic segmentation,''
  {\em arXiv preprint arXiv:2011.00147}, 2020.

\bibitem{chen2019crdoco}
Y.-C. Chen, Y.-Y. Lin, M.-H. Yang, and J.-B. Huang, ``Crdoco: Pixel-level
  domain transfer with cross-domain consistency,'' in {\em Proceedings of the
  IEEE/CVF Conference on Computer Vision and Pattern Recognition},
  pp.~1791--1800, 2019.

\bibitem{fu2019geometry}
H.~Fu, M.~Gong, C.~Wang, K.~Batmanghelich, K.~Zhang, and D.~Tao,
  ``Geometry-consistent generative adversarial networks for one-sided
  unsupervised domain mapping,'' in {\em Proceedings of the IEEE/CVF Conference
  on Computer Vision and Pattern Recognition}, pp.~2427--2436, 2019.

\bibitem{yang2020fda}
Y.~Yang and S.~Soatto, ``Fda: Fourier domain adaptation for semantic
  segmentation,'' in {\em Proceedings of the IEEE/CVF Conference on Computer
  Vision and Pattern Recognition}, pp.~4085--4095, 2020.

\bibitem{yang2020phase}
Y.~Yang, D.~Lao, G.~Sundaramoorthi, and S.~Soatto, ``Phase consistent
  ecological domain adaptation,'' in {\em Proceedings of the IEEE/CVF
  Conference on Computer Vision and Pattern Recognition}, pp.~9011--9020, 2020.

\bibitem{chen2018road}
Y.~Chen, W.~Li, and L.~Van~Gool, ``Road: Reality oriented adaptation for
  semantic segmentation of urban scenes,'' in {\em Proceedings of the IEEE
  Conference on Computer Vision and Pattern Recognition}, pp.~7892--7901, 2018.

\bibitem{zhang2019curriculum}
Y.~Zhang, P.~David, H.~Foroosh, and B.~Gong, ``A curriculum domain adaptation
  approach to the semantic segmentation of urban scenes,'' {\em IEEE
  transactions on pattern analysis and machine intelligence}, vol.~42, no.~8,
  pp.~1823--1841, 2019.

\bibitem{zou2018unsupervised}
Y.~Zou, Z.~Yu, B.~Kumar, and J.~Wang, ``Unsupervised domain adaptation for
  semantic segmentation via class-balanced self-training,'' in {\em Proceedings
  of the European conference on computer vision (ECCV)}, pp.~289--305, 2018.

\bibitem{pan2020unsupervised}
F.~Pan, I.~Shin, F.~Rameau, S.~Lee, and I.~S. Kweon, ``Unsupervised
  intra-domain adaptation for semantic segmentation through self-supervision,''
  in {\em Proceedings of the IEEE/CVF Conference on Computer Vision and Pattern
  Recognition}, pp.~3764--3773, 2020.

\bibitem{sun2019not}
R.~Sun, X.~Zhu, C.~Wu, C.~Huang, J.~Shi, and L.~Ma, ``Not all areas are equal:
  Transfer learning for semantic segmentation via hierarchical region
  selection,'' in {\em Proceedings of the IEEE/CVF Conference on Computer
  Vision and Pattern Recognition}, pp.~4360--4369, 2019.

\bibitem{li2020content}
G.~Li, G.~Kang, W.~Liu, Y.~Wei, and Y.~Yang, ``Content-consistent matching for
  domain adaptive semantic segmentation,'' in {\em European Conference on
  Computer Vision}, pp.~440--456, Springer, 2020.

\bibitem{zou2019confidence}
Y.~Zou, Z.~Yu, X.~Liu, B.~Kumar, and J.~Wang, ``Confidence regularized
  self-training,'' in {\em Proceedings of the IEEE International Conference on
  Computer Vision}, pp.~5982--5991, 2019.

\bibitem{wei2021theoretical}
C.~Wei, K.~Shen, Y.~Chen, and T.~Ma, ``Theoretical analysis of self-training
  with deep networks on unlabeled data,'' in {\em International Conference on
  Learning Representations}, 2021.

\bibitem{lian2019constructing}
Q.~Lian, F.~Lv, L.~Duan, and B.~Gong, ``Constructing self-motivated pyramid
  curriculums for cross-domain semantic segmentation: A non-adversarial
  approach,'' in {\em Proceedings of the IEEE/CVF International Conference on
  Computer Vision}, pp.~6758--6767, 2019.

\bibitem{mei2020instance}
K.~Mei, C.~Zhu, J.~Zou, and S.~Zhang, ``Instance adaptive self-training for
  unsupervised domain adaptation,'' in {\em Computer Vision--ECCV 2020: 16th
  European Conference, Glasgow, UK, August 23--28, 2020, Proceedings, Part XXVI
  16}, pp.~415--430, Springer, 2020.

\bibitem{tang2021unsupervised}
S.~Tang, P.~Tang, Y.~Gong, Z.~Ma, and M.~Xie, ``Unsupervised domain adaptation
  via coarse-to-fine feature alignment method using contrastive learning,''
  {\em arXiv preprint arXiv:2103.12371}, 2021.

\bibitem{zhang2021prototypical}
P.~Zhang, B.~Zhang, T.~Zhang, D.~Chen, Y.~Wang, and F.~Wen, ``Prototypical
  pseudo label denoising and target structure learning for domain adaptive
  semantic segmentation,'' in {\em Proceedings of the IEEE/CVF Conference on
  Computer Vision and Pattern Recognition}, pp.~12414--12424, 2021.

\bibitem{kundu2021generalize}
J.~N. Kundu, A.~Kulkarni, A.~Singh, V.~Jampani, and R.~V. Babu, ``Generalize
  then adapt: Source-free domain adaptive semantic segmentation,'' in {\em
  Proceedings of the IEEE/CVF International Conference on Computer Vision},
  pp.~7046--7056, 2021.

\bibitem{sun2019test}
Y.~Sun, X.~Wang, Z.~Liu, J.~Miller, A.~A. Efros, and M.~Hardt, ``Test-time
  training for out-of-distribution generalization,'' 2019.

\bibitem{ioffe2015batch}
S.~Ioffe and C.~Szegedy, ``Batch normalization: Accelerating deep network
  training by reducing internal covariate shift,'' in {\em International
  conference on machine learning}, pp.~448--456, PMLR, 2015.

\bibitem{li2016revisiting}
Y.~Li, N.~Wang, J.~Shi, J.~Liu, and X.~Hou, ``Revisiting batch normalization
  for practical domain adaptation,'' {\em arXiv preprint arXiv:1603.04779},
  2016.

\bibitem{chang2019domain}
W.-G. Chang, T.~You, S.~Seo, S.~Kwak, and B.~Han, ``Domain-specific batch
  normalization for unsupervised domain adaptation,'' in {\em Proceedings of
  the IEEE/CVF Conference on Computer Vision and Pattern Recognition},
  pp.~7354--7362, 2019.

\bibitem{nado2020evaluating}
Z.~Nado, S.~Padhy, D.~Sculley, A.~D'Amour, B.~Lakshminarayanan, and J.~Snoek,
  ``Evaluating prediction-time batch normalization for robustness under
  covariate shift,'' {\em arXiv preprint arXiv:2006.10963}, 2020.

\bibitem{frankle2021training}
J.~Frankle, D.~J. Schwab, and A.~S. Morcos, ``Training batchnorm and only
  batchnorm: On the expressive power of random features in {\{}cnn{\}}s,'' in
  {\em International Conference on Learning Representations}, 2021.

\bibitem{cubuk2020randaugment}
E.~D. Cubuk, B.~Zoph, J.~Shlens, and Q.~V. Le, ``Randaugment: Practical
  automated data augmentation with a reduced search space,'' in {\em
  Proceedings of the IEEE/CVF Conference on Computer Vision and Pattern
  Recognition Workshops}, pp.~702--703, 2020.

\bibitem{sajjadi2016regularization}
M.~Sajjadi, M.~Javanmardi, and T.~Tasdizen, ``Regularization with stochastic
  transformations and perturbations for deep semi-supervised learning,'' {\em
  arXiv preprint arXiv:1606.04586}, 2016.

\bibitem{berthelot2019mixmatch}
D.~Berthelot, N.~Carlini, I.~Goodfellow, N.~Papernot, A.~Oliver, and C.~A.
  Raffel, ``Mixmatch: A holistic approach to semi-supervised learning,'' in
  {\em Advances in Neural Information Processing Systems}, pp.~5049--5059,
  2019.

\bibitem{xie2020unsupervised}
Q.~Xie, Z.~Dai, E.~Hovy, M.-T. Luong, and Q.~V. Le, ``Unsupervised data
  augmentation for consistency training,'' {\em arXiv preprint
  arXiv:1904.12848}, 2020.

\bibitem{sohn2020fixmatch}
K.~Sohn, D.~Berthelot, C.-L. Li, Z.~Zhang, N.~Carlini, E.~D. Cubuk, A.~Kurakin,
  H.~Zhang, and C.~Raffel, ``Fixmatch: Simplifying semi-supervised learning
  with consistency and confidence,'' {\em arXiv preprint arXiv:2001.07685},
  2020.

\bibitem{li2020rethinking}
B.~Li, Y.~Wang, T.~Che, S.~Zhang, S.~Zhao, P.~Xu, W.~Zhou, Y.~Bengio, and
  K.~Keutzer, ``Rethinking distributional matching based domain adaptation,''
  {\em arXiv preprint arXiv:2006.13352}, 2020.

\bibitem{araslanov2021self}
N.~Araslanov and S.~Roth, ``Self-supervised augmentation consistency for
  adapting semantic segmentation,'' in {\em Proceedings of the IEEE/CVF
  Conference on Computer Vision and Pattern Recognition}, pp.~15384--15394,
  2021.

\bibitem{bahat2019natural}
Y.~Bahat, M.~Irani, and G.~Shakhnarovich, ``Natural and adversarial error
  detection using invariance to image transformations,'' {\em arXiv preprint
  arXiv:1902.00236}, 2019.

\bibitem{chen2019domain}
M.~Chen, H.~Xue, and D.~Cai, ``Domain adaptation for semantic segmentation with
  maximum squares loss,'' in {\em Proceedings of the IEEE/CVF International
  Conference on Computer Vision}, pp.~2090--2099, 2019.

\bibitem{ramos2003using}
J.~Ramos {\em et~al.}, ``Using tf-idf to determine word relevance in document
  queries,'' in {\em Proceedings of the first instructional conference on
  machine learning}, vol.~242, pp.~29--48, Citeseer, 2003.

\bibitem{wei2020theoretical}
C.~Wei, K.~Shen, Y.~Chen, and T.~Ma, ``Theoretical analysis of self-training
  with deep networks on unlabeled data,'' in {\em International Conference on
  Learning Representations}, 2020.

\bibitem{paszke2019pytorch}
A.~Paszke, S.~Gross, F.~Massa, A.~Lerer, J.~Bradbury, G.~Chanan, T.~Killeen,
  Z.~Lin, N.~Gimelshein, L.~Antiga, {\em et~al.}, ``Pytorch: An imperative
  style, high-performance deep learning library,'' {\em arXiv preprint
  arXiv:1912.01703}, 2019.

\bibitem{kingma2014adam}
D.~P. Kingma and J.~Ba, ``Adam: A method for stochastic optimization,'' {\em
  arXiv preprint arXiv:1412.6980}, 2014.

\bibitem{chen2017rethinking}
L.-C. Chen, G.~Papandreou, F.~Schroff, and H.~Adam, ``Rethinking atrous
  convolution for semantic image segmentation,'' {\em arXiv preprint
  arXiv:1706.05587}, 2017.

\bibitem{he2016deep}
K.~He, X.~Zhang, S.~Ren, and J.~Sun, ``Deep residual learning for image
  recognition,'' in {\em Proceedings of the IEEE conference on computer vision
  and pattern recognition}, pp.~770--778, 2016.

\bibitem{russakovsky2015imagenet}
O.~Russakovsky, J.~Deng, H.~Su, J.~Krause, S.~Satheesh, S.~Ma, Z.~Huang,
  A.~Karpathy, A.~Khosla, M.~Bernstein, {\em et~al.}, ``Imagenet large scale
  visual recognition challenge,'' {\em International journal of computer
  vision}, vol.~115, no.~3, pp.~211--252, 2015.

\bibitem{kang2019decoupling}
B.~Kang, S.~Xie, M.~Rohrbach, Z.~Yan, A.~Gordo, J.~Feng, and Y.~Kalantidis,
  ``Decoupling representation and classifier for long-tailed recognition,''
  {\em arXiv preprint arXiv:1910.09217}, 2019.

\bibitem{chen2017deeplab}
L.-C. Chen, G.~Papandreou, I.~Kokkinos, K.~Murphy, and A.~L. Yuille, ``Deeplab:
  Semantic image segmentation with deep convolutional nets, atrous convolution,
  and fully connected crfs,'' {\em IEEE transactions on pattern analysis and
  machine intelligence}, vol.~40, no.~4, pp.~834--848, 2017.

\end{thebibliography}
}

\localtableofcontents

\newcommand\DoToC{%
  \startcontents
  \printcontents{}{2}{\textbf{Contents}\vskip3pt\hrule\vskip5pt}
  \vskip3pt\hrule\vskip5pt
}

\section{$\method$: Additional analysis} 

\subsection{Noisy reliability oracles}

In Sec 4.2 of the main paper, we presented the performance for a reliability oracle designed as \method with perfect reliability maps (using ground truth labels). We found that this achieves an mIoU of 49.35 within a single epoch -- a 2.23 mIoU gain over our method, capturing the performance drop due to imperfect reliability maps. 

\begin{figure}[t]
  \centering
  \includegraphics[width=0.75\textwidth]{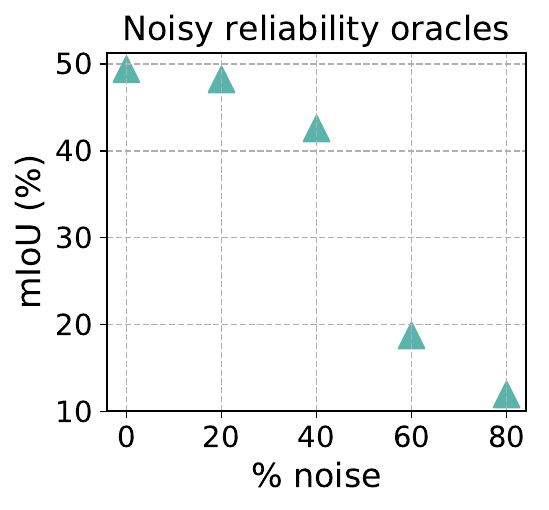}
  \caption{GTA$\rightarrow$Cityscapes: Reliability oracles with varying amount of noise.}
  \label{fig:noisy_oracle}
\end{figure}

In Fig.~\ref{fig:noisy_oracle} we extend this experiment to train reliability oracles with a varying amount of noise. Concretely, we begin with a perfect reliability mapping and randomly flip the mapping for P-\% of pixels in each image -- we vary P across $\{0, 20, 40, 60, 80\}$, where P$=0$ corresponds to the perfect reliability oracle reported in the main paper. As seen, performance is approximately the same upto P=20 but begins to dip at P=40, dropping off steeply afterwards to below-source mIoU values. Clearly, a good reliability measure is critical to the success of our method, and our combination of pixel-level consistency and confidence is an effective choice that achieves an mIoU of 47.12 (higher than the oracle with 40\% noise) despite only using self-supervised consistency and confidence signals.

\subsection{Source-equipped training.}

While \method is designed as a \emph{source-free} adaptation strategy, for completeness we also report the performance of a source-equipped version of it wherein we assume access to labeled source data and minimize an additional supervised cross-entropy loss. We observe an mIoU of 44.92 after one epoch of adaptation (performance worsens at later epochs), lower than in the source-free setting (47.12). We hypothesize that this is due to restricting updates to batch normalization parameters, as it is challenging to learn a shared set of batch-normalization parameters that perform well on both source and target data. To test this hypothesis, we try training all parameters in this source-equipped setting for 5 epochs with a learning rate of $5\times10^{-6}$. We find that performance steadily improves to achieve an mIoU of 47.78 after 5 epochs.

\subsection{Pseudolabel accuracy by type over time}

\begin{figure}[t]
  \centering
  \includegraphics[width=0.75\textwidth]{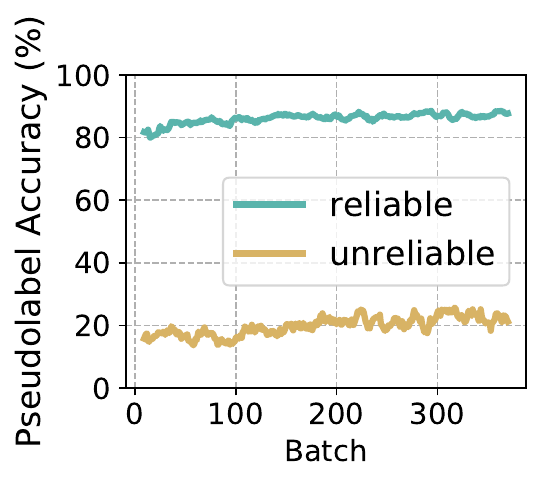}
  \caption{GTA$\rightarrow$Cityscapes: Pseudolabel accuracy over the course of training (1 epoch).}
  \label{fig:accvtime}
\end{figure}

Recall that in Sec. 4.3 of the main paper, we reported the aggregate accuracy of pseudolabels marked as reliable and unreliable, finding that reliable pseudolabels are indeed significantly more accurate (86.2\%) than unreliable ones (19.1\%). In Fig.~\ref{fig:accvtime}, we plot these accuracies for pseudolabels of different types as training progresses. 

As seen, we observe similar trends throughout the course of adaptation -- reliable pixels are indeed significantly more accurate than unreliable ones throughout the course of training, and their accuracy gradually increases over time as the model adapts to the target domain. Further, unreliable pixels are subtantially less accurate.

\subsection{Alternative combinations of consistency and confidence} 

Recall that we opt to select as reliable pixel predictions that are either consistent across predictive views or in the top-K \%-ile of model confidence for each category, and everything else as unreliable. As alternative combinations of consistency and confidence, we experiment with treating predictions that are consistent AND confident as reliable. After training in this setting, we observe a low mIoU of 22.16. Altogether, we find that consistency and confidence complement each other when used individually -- only selecting pixels that are both consistent AND confident is a highly restrictive criterion that leads to selecting a very small percentage of pixels for self-training (improving the precision with respect to correctness but at a large cost to the recall). Ultimately this leads to poor performance.

\section{Additional training details} 

\subsection{Code}

\noindent We will publicly release all our code and pretrained models.

\subsection{Implementation details}

\noindent \textbf{Augmentation}. For spatial augmentations, we use random cropping of an area 25-50\% that of the original image, followed by resizing back to the original resolution. We match aspect ratio to the original image (=2.0). For color jitter transformations, we use a subset of RandAugment~\cite{cubuk2020randaugment} transforms: specifically AutoContrast, Equalize, Brightness, and Sharpness. For each image, a single transform from this list is selected at random and applied with a severity of 2.0 (maximum severity=30). 

\noindent \textbf{DeepLabV2 training details}. We now provide implementation details for our experiments with DeepLabV2~\cite{chen2017deeplab} with a ResNet101~\cite{he2016deep} backbone. As before, we only update batch norm parameters and report performance after 1 epoch of adaptation. We use a learning rate of $2.5\times10^{-4}$ (GTA5$\to$Cityscapes) \& $5\times10^{-5}$ (SYNTHIA$\to$Cityscapes), no weight decay, the Adam~\cite{kingma2014adam} optimizer, and a batch size of 8. We match source-training exactly to Chen~\emph{et al.}~\cite{chen2019domain}.

\subsection{Dataset licenses} 

\noindent \textbf{GTA5:} Images in the GTA5 dataset were collected from the GTA5 game and are distributed under the MIT license which makes it free to use for research. 

\noindent \textbf{SYNTHIA:} SYNTHIA is distributed under a Creative Commons license which allows free use for non-commercial purposes.

\noindent \textbf{Cityscapes:} The Cityscapes dataset license allows free use for non-commercial purposes.

\noindent \textbf{Dark Zurich:} The Dark Zurich dataset is released under a Creative Commons license which allows free use for non-commerical purposes.

\noindent To the best of our knowledge, none of the above datasets contain personally identifiable information (in fact, GTA5 and SYNTHIA comprise of purely synthetic images), or offensive content. 

\subsection{Baselines}

\noindent We briefly describe the source-equipped baselines that we report in Sec. 4, and refer readers to the original work for more details.

\noindent i) \textbf{AdvEnt}~\cite{vu2019advent}: Performs conditional entropy minimization and adversarial entropy minimization of the model's predictions over unlabeled target data. 

\noindent ii) \textbf{AdaptSegNet~\cite{tsai2018learning}}: Performs adversarial learning in the output space for domain alignment. 

\noindent iii) \textbf{MaxSquares}~\cite{chen2019domain}: Proposes a ``maximum squares'' loss which leads to more balanced gradients when  self-training on poorly-classified target samples than traditional entropy minimization. 

\noindent iv) \textbf{IAST~\cite{mei2020instance}}: Proposes ``instance adaptive self-training'', which combines a pseudolabel generation strategy with an instance-adaptive selector, and employs a region-guided regularization strategy to smoothen pseudolabels.

\end{document}